%% file: main.tex
\documentclass[12pt,letterpaper,english]{article}

\usepackage[top=2.54cm, bottom=3.54cm, left=2.54cm, right=2.54cm]{geometry} 
\usepackage[utf8]{inputenc} 
\usepackage[T1]{fontenc} 
\usepackage[english]{babel} 
\usepackage{graphicx,float,subcaption,caption} 
\usepackage{amsmath,mathtools,amsfonts} 
\usepackage{parskip} 
\usepackage{lipsum} 
\usepackage[ddmmyyyy]{datetime} 
\usepackage{array,booktabs,tabularx} 
\usepackage{xcolor} 
\usepackage{fancyhdr} 
\usepackage[lastpage,user]{zref} 
\usepackage{multirow} 
\usepackage{setspace} 
\usepackage{microtype} 
\usepackage{hyperref} 
\usepackage{enumitem} %
\usepackage[nottoc,numbib]{tocbibind} 
\usepackage{systeme} 
\usepackage{csquotes}
\usepackage[T1]{fontenc}
\usepackage{lmodern}
\usepackage[style=numeric,natbib=true]{biblatex}
\AtEveryBibitem{%
  \clearfield{issn} 
  \clearfield{doi} 

  \ifentrytype{online}{}{
    \clearfield{url}
  }
}

\bibliography{references}
\usepackage{graphicx}
\usepackage{capt-of}
\usepackage{booktabs}
\usepackage{varwidth}
\usepackage{lscape}
\usepackage{tabularx}
\usepackage{xcolor}
\usepackage[normalem]{ulem}

\usepackage[export]{adjustbox}

\newcommand{\nomProjet}{Driving Datasets Literature Review}

\hypersetup{
	colorlinks=true, 
	linkcolor=teal,          	
	urlcolor=blue!80!black,
    citecolor=gray,
    bookmarksdepth=4,
    pdftitle={\nomProjet}}  

\numberwithin{equation}{section} 
\numberwithin{figure}{section}
\numberwithin{table}{section}
\addto\captionsenglish{} 

\newcommand\debSec{\hspace{0.5cm}}

\newcolumntype{Y}{>{\centering\arraybackslash}X}

\newcolumntype{P}[1]{>{\centering\arraybackslash}p{#1}}

\begin{document}

\input{PageTitre/pageTitre.tex}

\renewcommand{\listfigurename}{List of Figures} 
\pagenumbering{Roman}
\rfoot{Page \thepage}
{\hypersetup{linkcolor=black}
    \tableofcontents \newpage
    \listoffigures 
    \listoftables}
\newpage

\pagenumbering{arabic}
\rfoot{Page \thepage \ of \zpageref{LastPage}}

\addcontentsline{toc}{section}{Introduction} 
\input{P01/introduction.tex}

\newpage
\addcontentsline{toc}{section}{Sensors \& Hardware} 
\input{P02/sensors.tex}

\newpage
\addcontentsline{toc}{section}{Calibration \& Synchronization} 
\input{P03/calibration.tex}

\newpage
\addcontentsline{toc}{section}{Taks} 
\input{P04/applications.tex}

\newpage
\addcontentsline{toc}{section}{Taks} 
\input{P05/datasets.tex}

\section*{Acknowledgments}
This work was supported by NSERC CRD Grant "BRITE: Bus RapId Transit systEm" (511843) and Consortium InnovÉÉ.


\newpage
\addcontentsline{toc}{section}{References} 
\nocite{*} 
\printbibliography

\end{document}

%% file: PageTitre/pageTitre.tex
\begin{titlepage}
	\begin{onehalfspace}
		\begin{center}
		    \textsc{\LARGE  }\\
			[2cm]
			\textsc{\LARGE Driving Datasets Literature Review}\\
			[1cm]
			\textbf{\LARGE for}\\
			[1cm]
			\textsc{\LARGE BRITE: Bus RapId Transit systEm}\\
			[2.5cm]
			\textsc{\LARGE Summer Internship 2019}\\
			[2cm]
			\includegraphics[height=.5in]{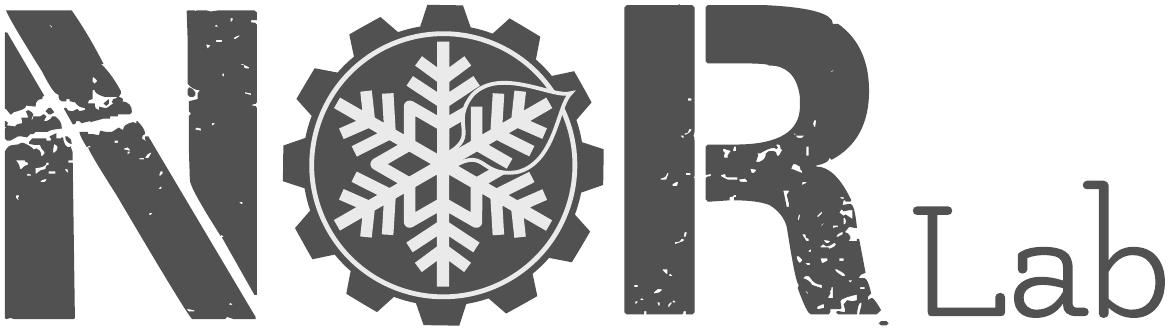} \\
			[0.5cm]
			\includegraphics[height=.5in]{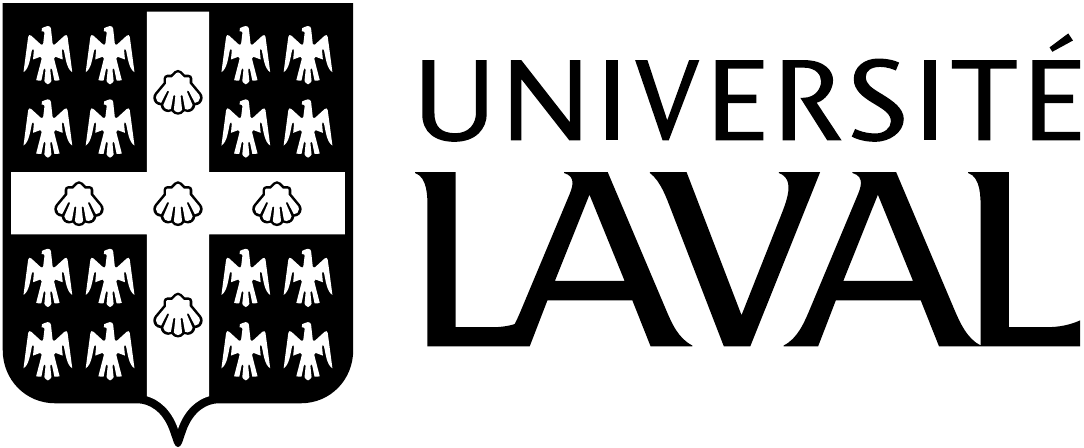} \\
			[2.5cm]
			\begin{flushleft}
				\textsc{\large
					Student: Charles-Éric Noël Laflamme}\\
				\textsc{\large
					Supervisor: Philippe Giguère}\quad \texttt{philippe.giguere@ift.ulaval.ca}\\
				\textsc{\large
					Cosupervisor: François Pomerleau}\quad
					\texttt{francois.pomerleau@ift.ulaval.ca}\\
				[0.5cm]
				\textsc{\large Laval University\\Québec City, Canada\\August $29^{th}$, 2019}
			\end{flushleft}
		
		\end{center}
	\end{onehalfspace}

\end{titlepage}
\leavevmode 

%% file: P01/introduction.tex
\section*{Introduction}\label{p01Sec:introduction}
\begin{spacing}{1.5}

\debSec For the last decade, the progress made in the autonomous driving scientific community and industry has been exceptional. 
With the rise of deep-learning and better hardware, algorithms embodying the different aspects of driving, such as lane following, obstacle detection, semantic segmentation, tracking, and motion estimation have reached unprecedented performance.
Although there are still no SAE Level-4 self-driving vehicles as of yet, recent developments in robotics and machine learning could soon make this aspiration a reality.

The availability of training data is a critical factor to the growth and success of autonomous driving. Although more powerful than traditional machine learning techniques, deep learning algorithms require a particularly massive amount of data for training and testing purposes. Moreover, in order to assimilate the entire driving process complexity and be reasonably safe, algorithms need to account for all possible real world scenarios, thus demanding highly dynamic and diverse datasets. Finally, it is often dangerous, costly and time-consuming to test driving algorithms on real vehicles.

This present document is a survey of the different autonomous driving datasets which have been published up to date. The first section introduces the many sensor types used in autonomous driving datasets. The second section investigates the calibration and synchronization procedure required to generate accurate data. The third section describes the diverse driving tasks explored by the datasets. Finally, the fourth section provides comprehensive lists of datasets, mainly in the form of tables.

\end{spacing}

%% file: P02/sensors.tex
\section{Sensors \& Hardware}
\label{p02Sec:sensors}
\begin{spacing}{1.5}

\debSec In order to achieve reliability and robustness, a wide variety of sensors are usually employed in autonomous vehicles. The diversity of sensing modalities also help mitigating difficult conditions, as their failure modes will be somewhat uncorrelated. These sensors can  be categorized into two main groups, namely \emph{exteroceptive} and \emph{proprioceptive} sensors.

\subsection{Exteroceptive Sensors}

Exteroceptive sensors are used to observe the environment, which in the case of autonomous vehicles means roads, buildings, cars, pedestrians, etc. The most common exteroceptive sensors for autonomous vehicles are cameras and range-sensing sensors.

\subsubsection{Cameras}
Cameras come in a variety of types and models. They are passive sensors meaning that they do not need to emit a signal to capture information, thereby limiting possible interference with other sensors. However, they are impacted negatively by illumination and weather conditions, due to their passive nature.

The most common type of camera is the monocular color camera. Being accessible, low-cost and straight-forward to use, monocular cameras have benefited from the majority of computer vision work of the last decades. Thus, most object detection, segmentation and tracking algorithms have been developed for these monocular cameras.

\begin{figure}[H]
	\centering
	\includegraphics[width=\textwidth]{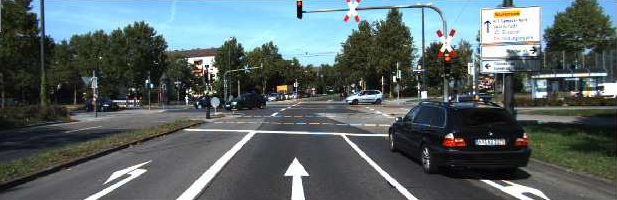}
	\caption[An example of monocular image from the KITTI datatset.]{An example of monocular image from the KITTI datatset \cite{Geiger2013IJRR}.}
	\label{fig:label}
\end{figure}

While driving, humans make use of their stereoscopic vision and focal distance information in order to judge depth and, for example, perform object avoidance. Human vision has also better resolution and a far wider field of view than most monocular cameras. In order to bridge the gap between 2D and 3D object detection and to gain more spatial information, monocular cameras are often used in stereo or multi-view systems. These setups are usually preferred because they present additional depth information, offer redundancy and provide a broader field of view. In order to be precise, these systems need to be calibrated methodically. A more detailed description of the calibration procedure can be found in Section~\ref{p03Sec:calibration}. Some manufacturers also offer precalibrated stereo camera systems, which can save time.

An alternative to arrays of cameras are omnidirectional cameras. These offer panoramic 360 degrees images, and are consequently often used to gain maximum information about a surrounding area. This can be highly beneficial for tasks such as localization and mapping. However, they tend to suffer heavily from lens distortion, which can affect the accuracy of a given task. Fish eye lens are also used in a similar fashion.

Other types of specialized cameras have been used in the past. For instance, to make up for the poor camera performance at night, thermal cameras and infrared cameras have been used for tasks such as pedestrian detection~\cite{hwang2015multispectral}.

\begin{figure}[H]
	\centering
	\includegraphics[width=\textwidth]{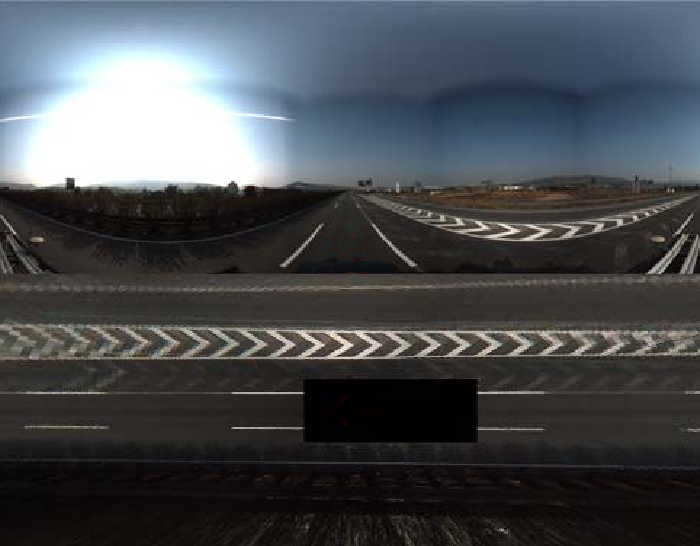}
	\caption[An example of an omnidirectional image.]{An example of an omnidirectional image from \cite{6932668}.}
	\label{fig:label}
\end{figure}

\begin{figure}[H]
	\centering
	\includegraphics[width=\textwidth]{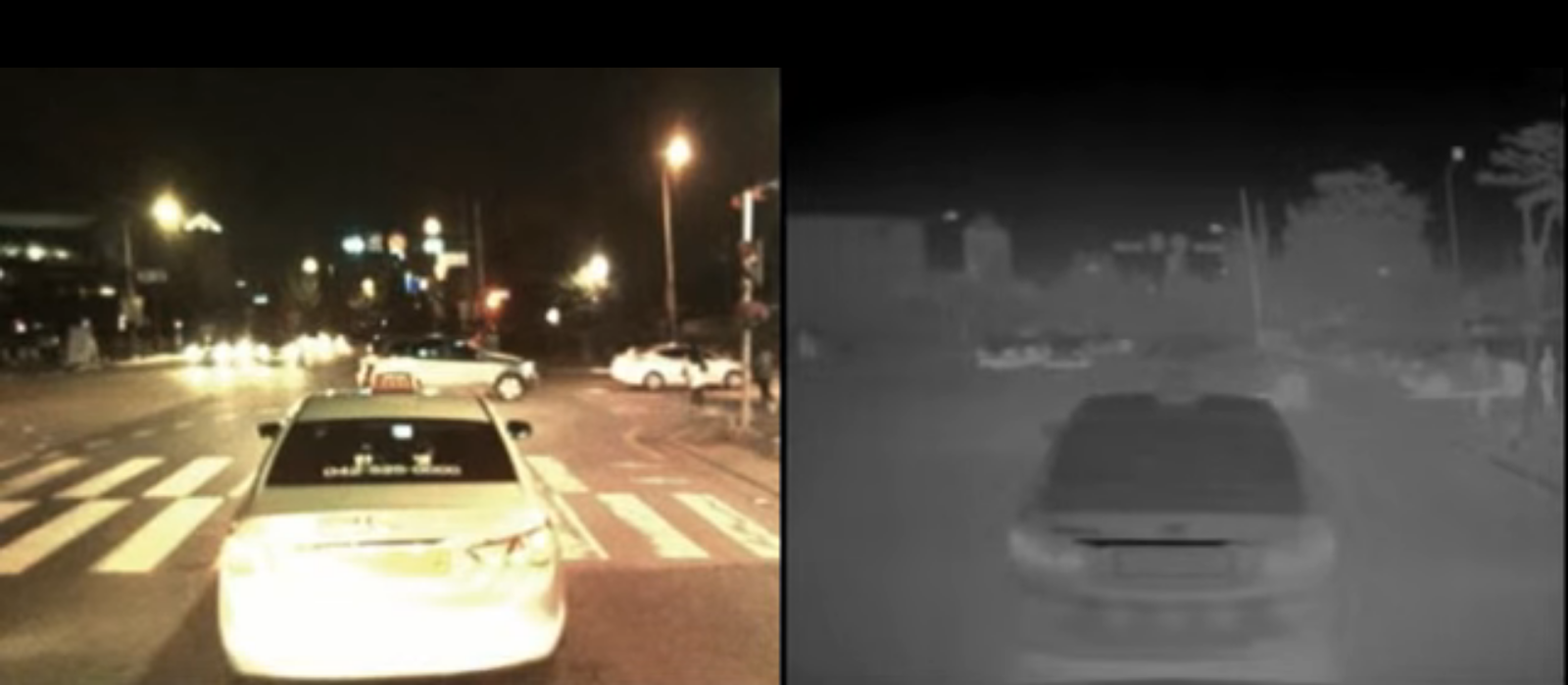}
	\caption[An example of a thermal image from the KAIST dataset.]{An example of a thermal image from the KAIST dataset \cite{hwang2015multispectral}.}
	\label{fig:label}
\end{figure}

Another type of cameras gaining interest are \emph{event cameras}, which output pixel-level brightness changes instead of standard intensity frames. They offer an excellent dynamic range and very low latency (in the order of $\mu$s), which can be quite useful in the case of highly-dynamic scenes. However, most already-developed vision algorithms cannot be readily applied to these cameras, as they output a sequence of asynchronous events rather than traditional intensity images. Nevertheless, some autonomous vehicle datasets with event cameras have now been published~\cite{DBLP:journals/corr/abs-1711-01458}. 

\begin{figure}[H]
	\centering
	\includegraphics[width=\textwidth]{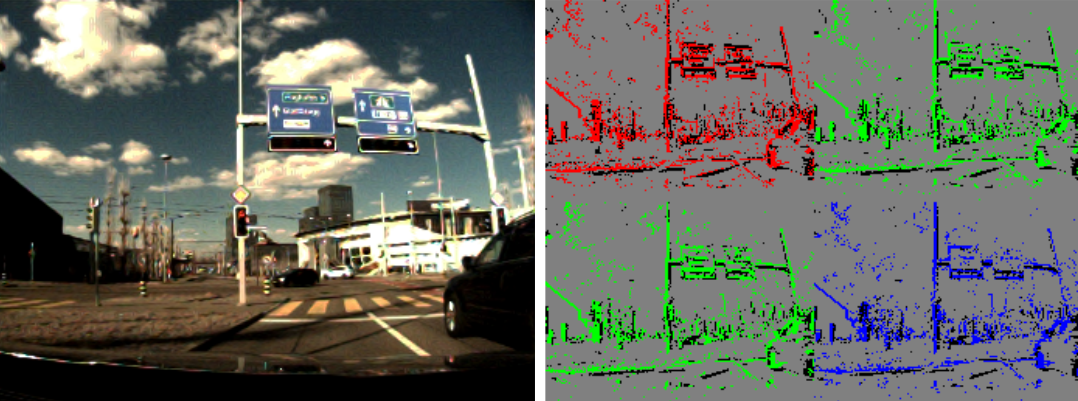}
	\caption[An example of an \emph{event image} from the DAVIS dataset.]{An example of an \emph{event image} from the DAVIS dataset~\cite{DBLP:journals/corr/abs-1711-01458}.}
	\label{fig:label}
\end{figure}

Finally, polarized sensors such as the \emph{Sony Pregius 5.0 MP IMX250} sensor have also recently reached better performance, which could potentially offer a higher level of detail. Polarization channels are often less affected by illumination changes and weather. They are also quite sensitive to a surface roughness, which could help with the detection of vehicles~\cite{Garcia:18}. However, no public autonomous driving datasets employing a polarization camera have been released, as of yet.

\begin{figure}[H]
	\centering
	\includegraphics[width=\textwidth]{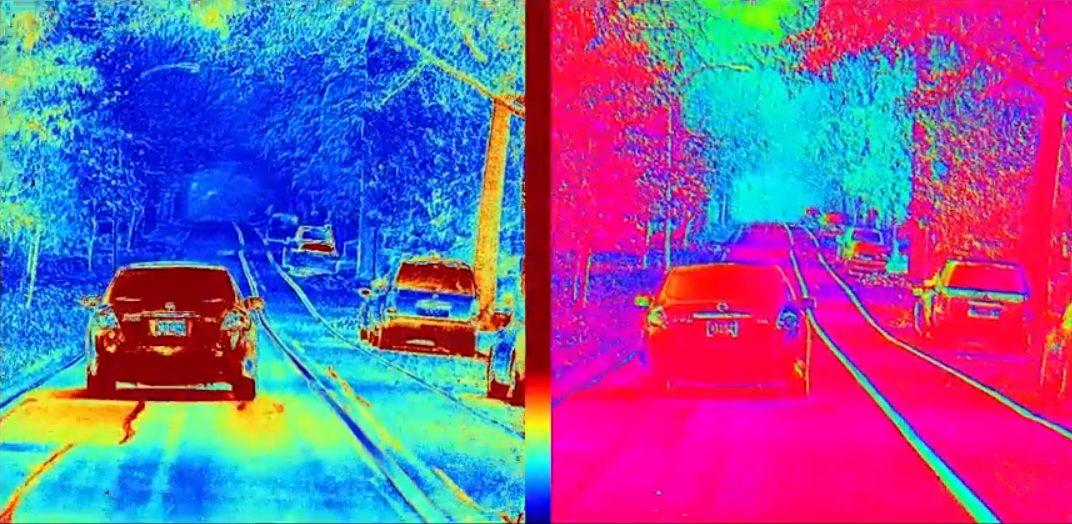}
	\caption[An example of a polarization image.]{An example of a polarization image from \cite{Garcia:18}.}
	\label{fig:label}
\end{figure}

\subsubsection{Range-sensing devices}

LiDARs, which stands for Light Detection and Ranging, detect objects and map their distances with great spatial coverage, in all lightning conditions. As such, they have been a sensor of choice for autonomous driving applications. 

The technology works by illuminating a target with an optical pulse, and measuring the characteristics of the reflected signal return in the same way a radar would detect reflected radio waves. They are much more accurate than radars, but their performance deteriorates from weather conditions such fog, rain or snow. They can also sometimes have trouble with detecting objects at close range.

LiDARs also come in a variety of formats, which can be split into two main families: \emph{i)}~mechanically spinning LiDARs and \emph{ii)}~solid-state LiDARs. While solid-state LiDARs are significantly cheaper, they suffer from a limited field-of-view compared to mechanical LiDARs. Solid-state LiDARs also tend to have a higher noise-to-signal ratio. They have also appeared much more recently than mechanical LiDARS. For all these reasons, most LiDARs used in autonomous datasets have been using either 2D or 3D mechanical LiDARs with often 360 degrees field-of-view. However, recent developments in solid state LiDARs are promising and they are slowly closing the performance gap.

It should be noted that LiDARs can scan a large amount of points at a very high rate, which can be a challenge for any algorithm run time.

In order to mitigate LiDAR limitations when it comes to adverse weather or close-range sensing, radars are also used as a range-sensing technology. Being a more mature sensor than LiDARs, radars are often much cheaper and lightweight, while also being able to determine the speed of its target. Nevertheless, they suffer from very poor spatial resolution, the difficulty of interpreting the received signals, and a much worse accuracy than LiDARs.

Finally, sonars are also used in the industry. While also cheap, sonars have very limited range and precision in addition to being susceptible to weather conditions. They are mainly used for nearby obstacle detection.

\subsection{Proprioceptive Sensors}

Proprioceptive sensors measure values internally to a given system. In the case of an autonomous vehicle, these measurements include linear and angular positions, speed and acceleration. Most modern cars are already equipped with a plethora of proprioceptive sensors. Wheel encoders are used for odometry, tachometers are used for speed, and IMUs can monitor acceleration changes. These sensors are often accessible through the vehicle's CAN bus.

However, the accuracy of sensors from car manufacturers are typically too low for autonomous vehicles applications, especially in the case of IMUs. For mapping purposes, the IMU measurements and odometry provide a point-matching algorithms such as Iterative Closest Point (ICP)~\cite{BeslMcKay} with an initial transformation guess, which is crucial to the algorithm performance (both in terms of speed and robustness).
In order to reach a higher level of precision, most autonomous vehicle datasets use a navigation-grade IMU along with a GPS.
Autonomous datasets dedicated to mapping and localization also often use an RTK GPS, which can provide centimeter-level accuracy, in order to compare the localization algorithms to a ground truth.

Other signals from the CAN bus protocol of a vehicle can be accessed, such as the steering angle, and the position of the accelerator and brake pedals. Such signals have been used by end-to-end learning algorithms~\cite{DBLP:journals/corr/abs-1711-01458,1812.05752}.

\end{spacing}

%% file: P03/calibration.tex
\section{Calibration \& Synchronization}
\label{p03Sec:calibration}
\begin{spacing}{1.5}

\debSec In order to achieve coherent data alignment, every sensor needs to be calibrated and synchronized. Below we describe both spatial alignment (calibration) and temporal alignment (synchronization).

\subsection{Calibration}
Calibration usually refers to a spatial-referencing process by which the relative coordinate frames of all sensors are established. For cameras, the calibration is essential to accurately measure object and distances on a scene for stereo camera setups. 
Camera calibration or \textit{camera resectioning} is often split into \emph{intrinsic} and \emph{extrinsic} parameters retrieval. Intrinsic parameters refer to the camera's inherent parameters such as focal length, principal point coordinates and distortion coefficients, to be used in image rectification. On the other hand, extrinsic parameters denote the coordinate system transformations from 3D world coordinates to 3D camera coordinates. These parameters are retrieved using referenced calibration points, called fiducial markers, usually with a checker board target with known dimensions~\cite{Zhang:2000:FNT:357014.357025}.

Once camera-to-camera calibration is achieved, stereoscopic depth reconstruction can be performed for all overlapping pixels. The additional depth channel can then be exploited by various machine learning algorithms to seize additional object features which can lead to better results. 

The depth channel derived from stereoscopic reconstruction can be as dense as the camera, depending on the texture in the scene. However, it can be strongly affected by the level of illumination in the environment. It is also far less precise at longer ranges, with the distance $z$ accuracy decreasing in $1/z^2$. Range sensors, such as LiDARs, are often far more precise and reliable at estimating object distances, but their measurements significantly less dense than cameras.

In a similar fashion, camera-to-range calibration can be achieved using reference plane surfaces~\cite{ICRA12}. However, the LiDAR-camera measurements fusion is not as straight-forward as stereoscopic reconstruction, because LiDAR data is a lot sparser than camera images. In order to create a depth mask, a LiDAR point cloud needs to be projected on the image plane and then upsampled.
It is also possible to instead project the RGB channels onto the LiDAR point cloud to generate a \emph{colored point cloud}. This approach is generally used by mapping and localization tasks, by providing extra information to a point-matching algorithm.

Finally, motion-sensing devices are also calibrated with each sensor using hand-eye calibration~\cite{doi:10.1177/027836499501400301} in order to reference the measurements to the inertial navigation system (INS). This is particularly important for localization and mapping tasks such as SLAM, where the inertial and odometry measurements are used as a first estimate for the new position and orientation, which is critical to these algorithms' performance.


\subsection{Synchronization}
While calibration addresses the spatial alignment of sensors, synchronization temporally matches measurements together. Driving is a highly-dynamic process within a rapidly-changing environment, making the synchronization process critical for data temporal alignment. In order to synchronize these different measurements, sensors are often triggered externally. The measurements of each sensor are also timestamped with a system clock. When sensors have different acquisition rates, and thus different timestamps, the measurements can either be interpolated or the closest measurement can be selected, depending on the use case.

The exposure time of a camera is nearly instantaneous and should not yield bad data alignment. However, most rotatory LiDARs execute a full rotation in about 0.1 $s$. This scanning speed is considerably slower than the car speed, which can introduce distortions in the point clouds. Techniques have been developed in order to account for the vehicle's motion within a LiDAR scan~\cite{article_lidar_motion}.

\end{spacing}

%% file: P04/applications.tex
\section{Tasks}\label{p04Sec:tasks}
\begin{spacing}{1.5}

\debSec Most autonomous driving approaches up to date have tried to deconstruct the complex driving process into different smaller and simpler sub-tasks. With this modular approach in mind, each dataset generated for the autonomous vehicle community have more often than not revolved around one or many of these specific tasks. Below, we enumerate the most popular ones.

\subsection{Stereo Vision}
As mentioned before, driving algorithms can benefit from having additional depth information. One of the simplest ways to acquire such 3D information is through stereo vision. Stereo vision is the task in which the depth of a scene is triangulated by identifying common features in two images taken from cameras mounted next to each other. In the case of driving, challenges for stereo vision include reflective and shiny surfaces such as car bodies. Repetitive structures like fences and transparent surfaces (glass) are other common failure cases.

Datasets dedicated to 3D reconstruction usually offer pixel-wise depth maps as a ground truth. Often, they have been generated by interpolating 3D LiDAR point clouds and by fitting 3D CAD models onto individual objects.

\begin{figure}[H]
	\centering
	\includegraphics[width=\textwidth]{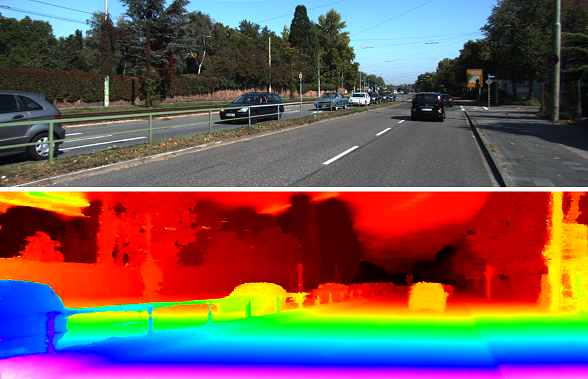}
	\caption[A sample depth map as ground truth for stereo vision, from the KITTI dataset]{A sample depth map as ground truth for stereo vision, from the KITTI dataset~\cite{DBLP:journals/corr/abs-1803-09719}.}
	\label{fig:label}
\end{figure}

It should be noted that with the rise of recent improvements in deep learning, monocular depth evaluation has also taken interest. In this case, the depth map is estimated using contextual information from a single image~\cite{7785097}.
Readers are referred to \textcite{gabr20183d} for a complete survey on stereo vision and other 3D reconstruction algorithms.

\subsection{Motion Estimation}

Because driving involves multiple objects moving at high speed, capturing the motion of objects in an image might yield desirable information. Optical flow, defined as finding the motion at each image location between consecutive frames, is one way of representing motion in a dense manner. Optical flow thus extracts additional motion information, which can be of particular importance for other tasks such as localization, ego-motion and tracking. 

Optical flow is restricted to monocular 2D images, which makes the retrieval of 3D motion challenging. \emph{Scene flow} is therefore defined as a generalized version of optical flow, where frames of stereo or multi-camera setups are used to establish motion.

Datasets dedicated to optical and scene flow usually offer optical flow fields, where a vector describing the motion for each pixel in the next or previous frame is provided.
Retrieving ground truth for both optical flow and scene flow is a time-consuming and tedious process and is often done by matching the image objects to 3D LiDAR maps.

\begin{figure}[H]
	\centering
	\includegraphics[width=\textwidth]{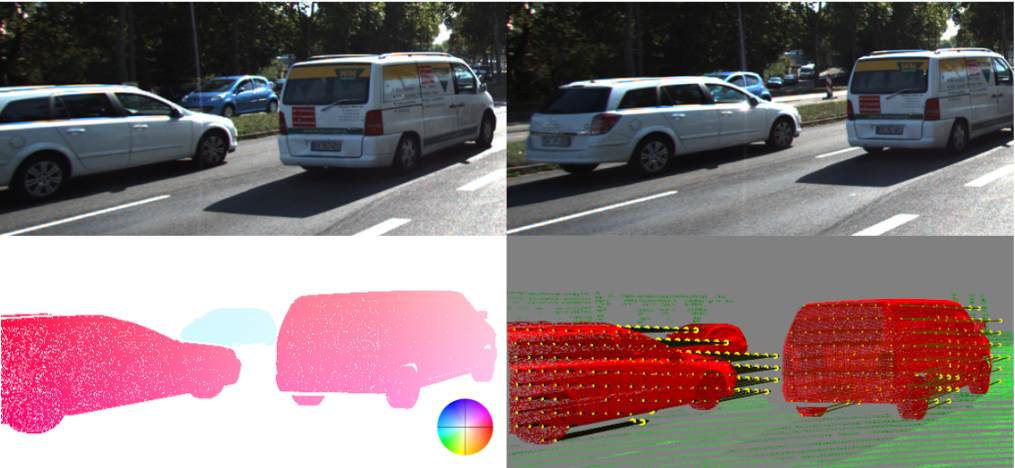}
	\caption[A sample of ground truth for optical flow (bottom left) and scene flow (bottom) right from the KITTI dataset]{A sample of ground truth for optical flow (bottom left) and scene flow (bottom) right from the KITTI dataset~\cite{DBLP:journals/corr/abs-1801-04720}.}
	\label{fig:label}
\end{figure}

An extensive survey on the state-of-the-art algorithms for optical and scene flow can be consulted in \cite{DBLP:journals/corr/YanX16}.

\subsection{Object Detection}
One of the first and foremost aspects of driving is the awareness of its surrounding. Whether it be pedestrians, other vehicles, traffic signs or obstacles, the detection and recognition of different objects in a scene is crucial to the safety and smooth functioning of an autonomous vehicle. Object detection addresses this task by determining the presence and localization of different predefined classes of objects in a scene.

Being an important and well-defined task, object detection has benefited from a considerable amount of attention in the computer vision and autonomous vehicle community. However, object detection still faces challenges. This is notably because of the wide variety of objects, weather conditions and illumination in a driving scene, along with heavy occlusion and truncation of objects~\cite{article-are-equal}.

Object detection itself can be split into subcategories depending  which modality is used to detect object, or what object itself is to be detected.

\begin{figure}[H]
	\centering
	\includegraphics[width=0.75\textwidth]{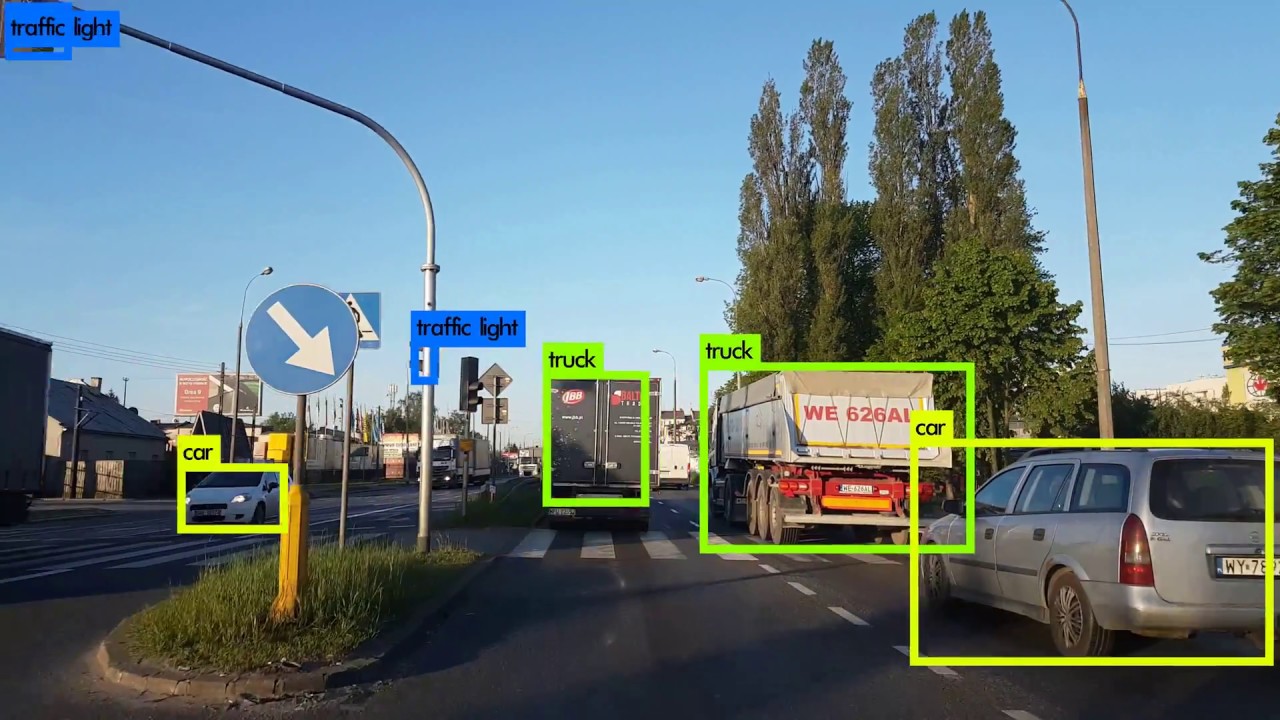}
	\caption[2D detection results from the YOLO algorithm]{2D detection results from the YOLO algorithm~\cite{yolov3}.}
	\label{fig:2D_detection}
\end{figure}

\begin{figure}[H]
	\centering
	\includegraphics[width=0.75\textwidth]{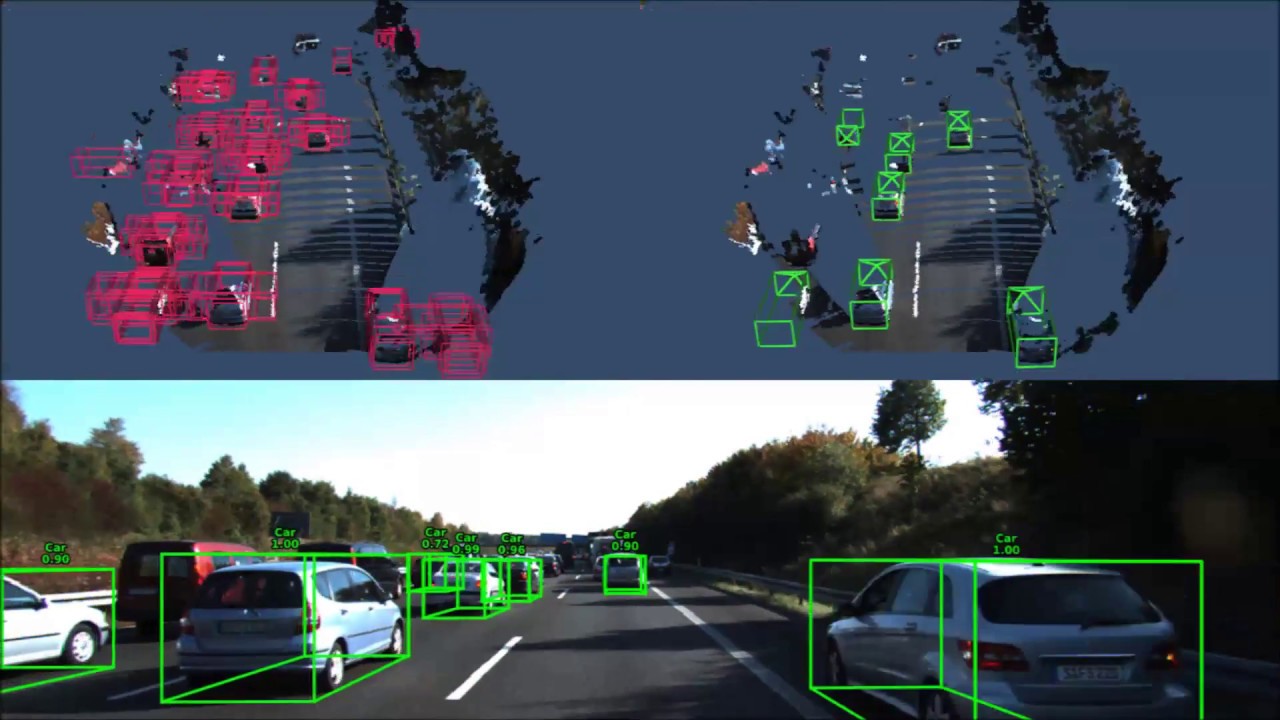}
	\caption[A sample ground truth frame for 3D object detection from the KITTI dataset]{A sample ground truth frame for 3D object detection from the KITTI dataset \cite{Geiger2013IJRR}.}
	\label{fig:3D_detection}
\end{figure}

Most object detection is done strictly on 2D images, hence the name 2D object detection. Each object is localized within the image, in pixel coordinates, as illustrated in \autoref{fig:2D_detection}. However, it should be noted that recent approaches have tried to include 3D features from either point cloud data~\cite{DBLP:journals/corr/ChenMWLX16} or stereo reconstruction~\cite{DBLP:journals/corr/ChenKZMFU16} in order to generate a more robust detection. Moreover, it is also possible to localize objects relative to the vehicle position in 3D space. This is commonly referred as 3D object detection, and is depicted in \autoref{fig:3D_detection}.

Datasets dedicated to object detection usually contain annotated data frames with 2D or 3D bounding boxes, which encloses the different objects, as ground truth.

Extensive reviews of deep learning detection techniques can be found in~\cite{DBLP:journals/corr/abs-1902-07830}, \cite{7066891}, \cite{DBLP:journals/corr/abs-1809-02165} and \cite{DBLP:journals/corr/abs-1905-05055}.

\subsection{Tracking}
Driving is a dynamic process with high-speed moving objects. Therefore, object detection is often insufficient in order to avoid collisions during path planning. Driving algorithms should not only predict the location of objects in a scene, but also their velocity and acceleration. In order to do so, tracking algorithms are used, which try to predict future positions of multiple moving objects based on the history of the individual positions.

\begin{figure}[H]
	\centering
	\includegraphics[width=0.75\textwidth]{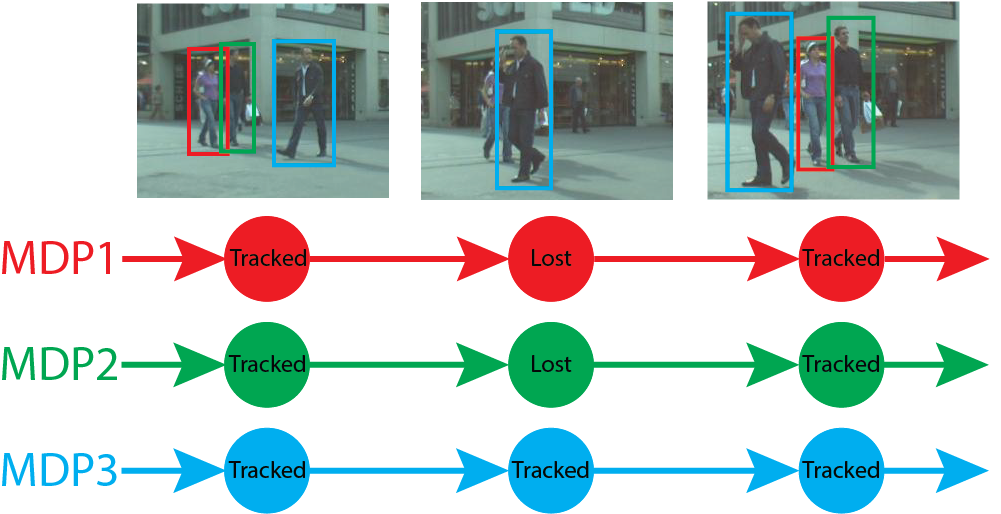}
	\caption[An example of tracking and occlusion challenges]{An example of tracking and occlusion challenges from \cite{7410891}.}
	\label{fig:Tracking}
\end{figure}

A popular and intuitive approach to tracking is \emph{tracking-by-detection}. An object-detection algorithm is first used to detect targets in each frame, which then need to be associated with each other over multiple frames. While efficient, this approach however suffers from detection errors and from the inherent difficulties of performing data-association. Tracking can also suffer if objects are momentarily occluded, as illustrated in \autoref{fig:Tracking}. It should be noted that pedestrian tracking is of particular interest, as they are the most vulnerable users of the road~\cite{dollarCVPR09peds}.

In order to predict a target trajectory in 3D, range information is most certainly needed for tracking. As mentioned before, such information can be obtained either by 3D reconstruction from cameras or from LiDAR point clouds, alone or through some sensor fusion process.

Just like object detection, tracking datasets usually contain annotations of data frames in the form of bounding boxes and labels, which are coherent over multiple frames, as a ground truth.

A thorough review on state-of-the-art tracking techniques is available in \textcite{article-tracking}.

\subsection{Semantic Segmentation}

Some objects such as roads, sidewalks and traffic lines are not well-defined by bounding boxes. Consequently, they need a more flexible representation, often down at the pixel-level. This problem is referred to as \emph{semantic segmentation}. 

\begin{figure}[H]
	\centering
	\includegraphics[width=\textwidth]{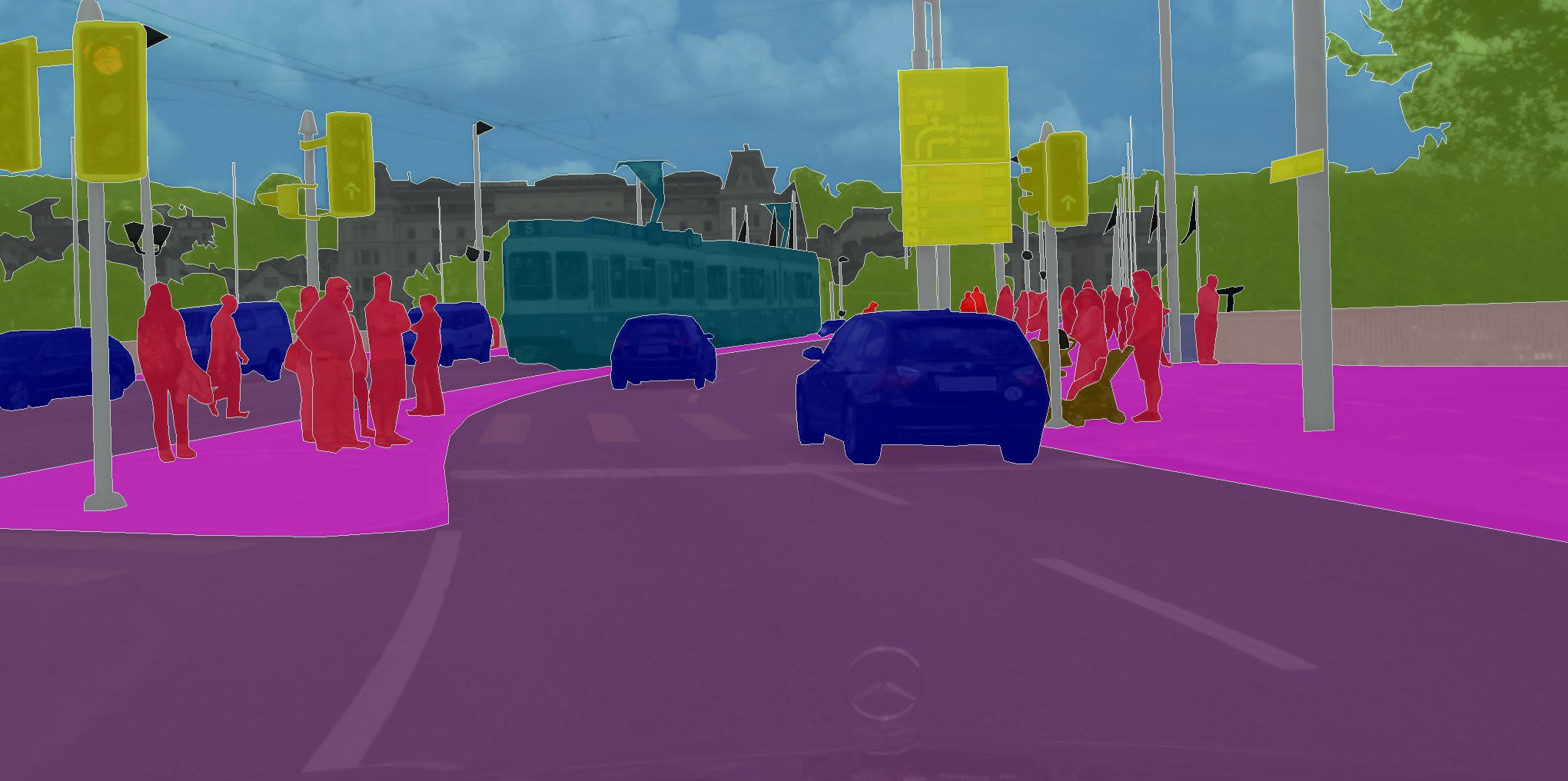}
	\caption[A sample ground truth frame for semantic segmentation from the Cityscapes dataset]{A sample ground truth frame for semantic segmentation from the Cityscapes dataset \cite{Cordts2016Cityscapes}.}
	\label{fig:Cityscape}
\end{figure}

Semantic segmentation is indeed similar to object detection in the way that it tries to locate different predefined classes of objects in a scene. However, instead of using bounding boxes to localize objects, each pixel of an image is assigned to a class, as seen in \autoref{fig:Cityscape}. The segmentation mask therefore offers a more dense and complex classification and localization, which can provide a better understanding of the scene. Semantic segmentation faces the same challenges as object detection such as occlusion, truncation and shadows, but also requires more complex computation. However, with model compression, pruning and hardware acceleration, it can reach real-time execution~\cite{Siam_2018_CVPR_Workshops}.

A more refined version of semantic segmentation is \emph{instance segmentation}, which not only classifies each pixel in a class, but also separates instances of the same class. Unlike semantic segmentation, instance segmentation thus provides information about each instance such as shape and position. Instance segmentation is particularly important to assess the trajectory of individual objects, for example vehicles, cyclists or pedestrians.

It should also be noted that in the case of driving, algorithms usually have access to multiple time frames of data. Methods which impose algorithms to be temporally coherent can improve segmentation accuracy and robustness.

Just like object detection, most of the previous work dedicated to segmentation has been done strictly on 2D images. However, shape and size are important features which cannot be exploited in the 2D space.To capture such information, LIDARs can of course be used.

\begin{figure}[H]
	\centering
	\includegraphics[width=\textwidth]{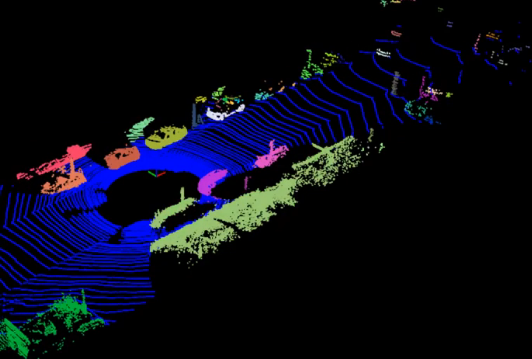}
	\caption[A sample ground truth frame for 3D instance segmentation]{A sample ground truth frame for 3D instance segmentation from \textcite{7989591}}
	\label{fig:3D_PC_labels}
\end{figure}

It it also possible to train semantic segmentation models strictly on point cloud data. However, generating accurate point cloud labels, such as the one displayed in \autoref{fig:3D_PC_labels}, is a tedious and time consuming task.

Given the importance of identifying drivable spaces, road and lane segmentation is of particular interest for autonomous vehicles. Along with spatially segmenting the road and lane itself, some algorithms also try to establish the host and neighbor lanes along with their direction. This information is particularly useful for tasks such as lane keeping, merging and turning.

Datasets dedicated to semantic segmentation usually annotate data frames with pixel-wise segmentation masks as ground truth, or in the case of 3D segmentation, voxel-wise masks.

\begin{figure}[H]
	\centering
	\includegraphics[width=\textwidth]{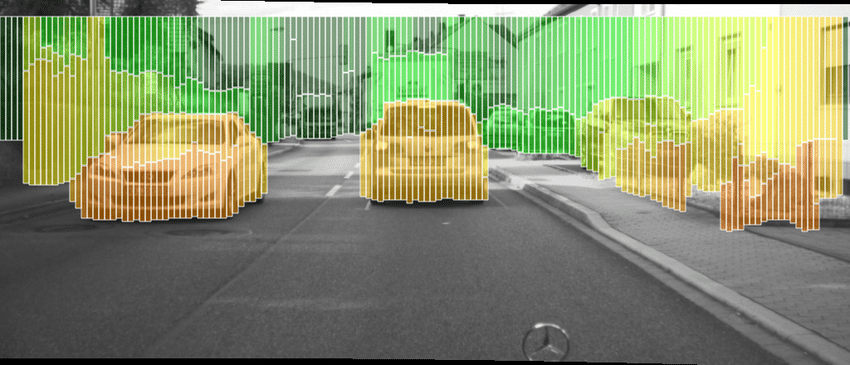}
	\caption[A sample ground truth frame for stixel segmentation from the Stixel dataset]{A sample ground truth frame for stixel segmentation from the Stixel dataset~\cite{DBLP:journals/corr/CordtsRSPERPF17}.}
	\label{fig:stixels}
\end{figure}

In order to alleviate computational burden of semantic segmentation, the \emph{stixel} representation has been suggested \cite{DBLP:journals/corr/CordtsRSPERPF17}. Stixels create a medium-level model of the environment, compressing pixel-wise information into vertical strips. An example of a stixel segmentation can be seen in \autoref{fig:stixels}.

For more details, a review on deep-learning techniques for semantic segmentation can be found in \textcite{DBLP:journals/corr/Garcia-GarciaOO17}. An in-depth survey on road and lane detection can be found in \textcite{BarHillel2014}.

\subsection{Localization}
Localization is a task critical to any mobile robot. In order to lay the appropriate path planning, a vehicle needs to know where it is exactly regarding its environment. Different approaches have been used in the past for localization.

The most straightforward one is the use of GPS and IMU sensors. While this combination of sensors is the most accessible and low-cost approach, it lacks the requirements needed for autonomous driving. Even with dead reckoning estimation from the IMU, commercial-grade GPS are simply too inaccurate. While RTK technology offers the precision needed for autonomous driving, the accuracy of the signal is highly dependant on the environment, urban setting with high buildings being particularly prone to errors from interference.

Simultaneous localization and mapping (SLAM) is another popular approach. It tries to generate a map on the fly using a vehicle's sensors, while estimating at the same time the position of the vehicle in the constructed map. It has the advantage of not needing any prior information about the environment, meaning this approach can work in any setting. However, SLAM still faces challenges as it is computationally heavy and needs to handle large-scale environments in real-time. Moreover, SLAM is prone to diverge in difficult environments. RTKs positioning is often used as ground truth for SLAM, given an appropriate signal reception.

Using pre-constructed maps is an alternative to SLAM that alleviates the problem on generating a map on the fly. Using a point-matching algorithm or visual landmark searches approach, a priori map-based localization algorithms can be highly accurate. However, a major weakness of these approaches is the fact that roads themselves are not completely static, and therefore the maps used for localization have to be updated for construction work or weather changes.

For further details, a complete survey on state-of-the-art techniques for localization can be found in \textcite{8306879}.

\subsection{Behaviour Analysis}

If driving vehicles are one day a reality, they will most likely have to interact with humans. Whether it be infering a pedestrian's intention to cross a street, identifying a driver's intent to perform a certain action or spotting potentially reckless maneuvers, autonomous vehicles need to have a high-level understanding of surrounding human behavior. The assessment of human behavior is therefore paramount for any autonomous driving applications.

While the task of behavior assessment is not as bounded as previously described tasks, recent datasets have tried to capture such human behavior. For instance, some datasets \cite{DBLP:journals/corr/JainKSRSS16,7293673,DBLP:journals/corr/abs-1811-02307} have tried annotating each of the driver's actions. Such data can be used to develop action-predicting algorithms, which can then be used to assess if a driver's maneuver is completely safe or not. 

\begin{figure}[H]
	\centering
	\includegraphics[width=0.75\textwidth]{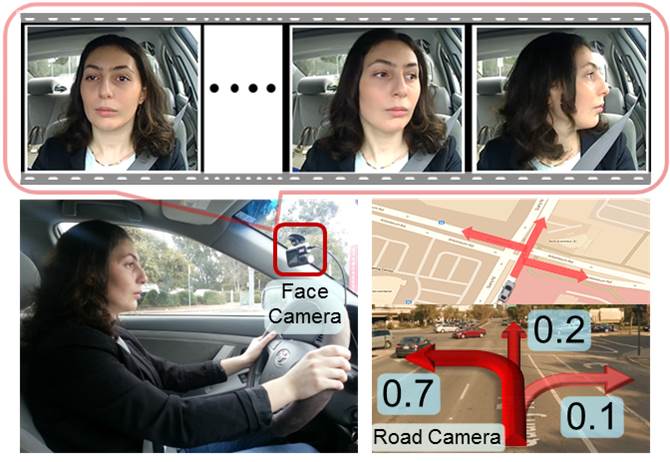}
	\caption[An example of driver face monitoring and predicted maneuvers from the Brain4Cars dataset]{An example of driver face monitoring and predicted maneuvers from the Brain4Cars dataset \cite{DBLP:journals/corr/JainKSRSS16}.}
	\label{fig:FaceMonitoring}
\end{figure}

Driver face monitoring is also a modality often used in order to predict maneuvers or visual focus, as shown in \autoref{fig:FaceMonitoring}. Some datasets even project the driver's gaze onto the road image, in order to know exactly on what the driver is focusing~\cite{DBLP:journals/corr/PalazziACSC17}. Such information can be used to established the driver's attention level for a safer driving experience.

Another important behavior assessment task is driving style recognition. Driving style can be defined in various ways including fuel consumption, brake-use, distance-keeping and aggressiveness. Establishing a driver's style can be used to adjust driving strategy, such as lane merging or alert the driver if he is being reckless~\cite{inproceedingsdrivesafe}.

\begin{figure}[H]
	\centering
	\includegraphics[width=0.75\textwidth]{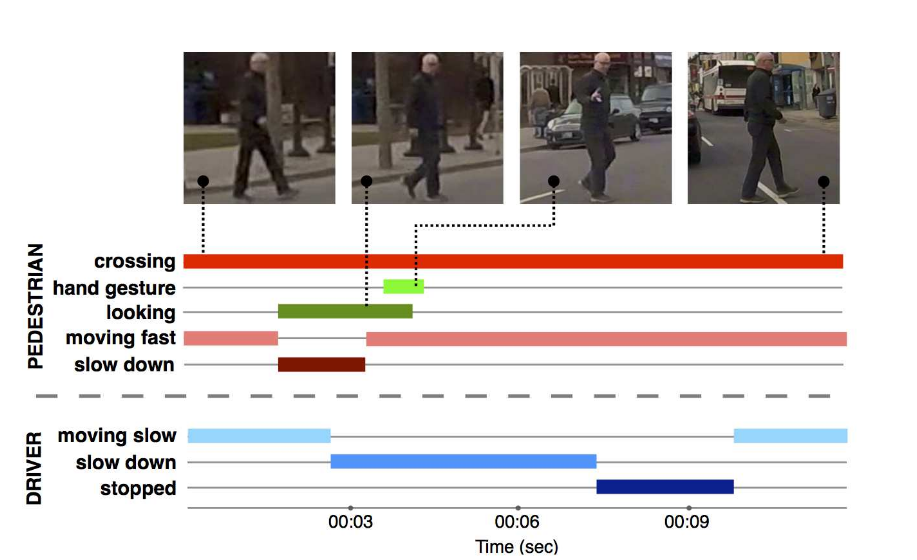}
	\caption[An example of labelled pedestrian and driver intention sequence from the JAAD dataset]{An example of labelled pedestrian and driver intention sequence from the JAAD dataset \cite{8265243}.}
	\label{fig:PedestrianBehavior}
\end{figure}

Finally, some datasets~\cite{DBLP:journals/corr/KotserubaRT16} have also collected data regarding pedestrian intention, as displayed in \autoref{fig:PedestrianBehavior}. Algorithms can then be trained to recognize whether a pedestrian wants to cross a street or not, and help prevent collisions.

For a more complete literature review on driving style recognition, pedestrian autonomous vehicle interactions or intersection behavior, readers are invited to consult \textcite{8002632}, \textcite{8667866} and \textcite{7556973} respectively.
  

\end{spacing}

%% file: P05/datasets.tex
\section{Datasets}
\label{sec:datasets}
\begin{spacing}{1.5}
\debSec
The following section presents the available open driving datasets, sorted by their respective task. It should be noted that only datasets which are in a real-driving context were selected. Therefore, synthetic datasets are not presented. Moreover, datasets acquired from terrestrial vehicles which are not cars, such as Segways or another robotic platform are also ignored. Finally, stationary-acquired datasets are also ignored.

\subsection{Object detection}
The \autoref{tab:object} table presents the autonomous driving datasets available for object detection and tracking, ordered chronologically. First of all, the shear size of the table is a great testimony to the level of attention the field has received in the autonomous driving community. Indeed, most datasets in this table have been published during the past three years, demonstrating that this is an active field. The table also attests the recognition autonomous driving has gained in the industry, with most of the major and recent datasets being published by companies such as Waymo, Aptiv, Lyft, Bosch and Hesai.

Moreover, by ordering the datasets by year, several interesting trends can be observed. The number of annotated frames per dataset seems to have increased considerably throughout the year. This phenomenon can probably be explained by the surge of deep learning in computer vision, which demands a vast amount of data. Furthermore, deep learning also gain from data diversity, which can explain why recent datasets seem to encompass more driving situations such as weather, time and traffic variety.  

The interest in multi-modal learning and the democratization of LiDAR sensors can also be observed in this table, with most of the recent datasets incorporating LiDARs and other sensing devices. The annotations have also been refined in the recent years, with the arrival of more annotated classes and 3D bounding boxes.

Finally, the table also presents object detection datasets which are aimed at more specialized tasks such as pedestrian detection or traffic sign detection. However, these kinds of niche datasets seem to be less frequent nowadays, which suggests that such specialized task have become trivially easy for object detection algorithms. It also seems that the state-of-the-art emphasizes rather on object detection generalization.

\subsection{Object segmentation}
Likewise, \autoref{tab:segementation} presents the driving datasets aimed at semantic and instance segmentation. While object detection and tracking are analogous to semantic and instance segmentation, there seems to be considerably less datasets dedicated to the latter tasks. This can presumably be attributed to the fact that generating segmentation annotation is a time-consuming and costly process.

Nonetheless, comparable trends can be observed from the segmentation datasets. Recent datasets are bigger and more diverse. The use of LiDARs and point cloud annotations are also very recent. Finally, most recent datasets are, again, published by companies.

\subsection{Lane detection}
The \autoref{tab:lane} presents lane detection datasets, another important part of vehicle perception. It can be thought as a special case of object detection or semantic segmentation. In fact, the consensus on the annotation type among the autonomous driving community does not seem to be established. While most datasets use spline lines to describe lanes, other datasets use pixel-wise annotations, bounding boxes or point cloud annotations.

It should be noted that while recent datasets have become larger and more exhaustive, there seems to be no datasets explicitly exploring damaged lane markings and lane-detection fail cases.

\subsection{Optical flow}
The list of optical flow datasets presented in \autoref{tab:optical} is noticeably shorter than for the previous tasks. The lack of driving optical flow datasets can most probably be explained by two causes. First, measuring ground truth for optical flow is a complicated and precise task which can be very hard to do in a highly-chaotic environment such as driving. This is why most benchmarked optical flow datasets are usually done in a controlled environment~\cite{Baker2011} or using synthetic data~\cite{Butler:ECCV:2012}. Secondly, most modern computer vision algorithms do not make use of optical flow data and thus the field has been losing interest over the years.
Nevertheless, the KITTI FLOW 2015~\cite{7298925} dataset offers a precise optical flow benchmark, with 3D fitted CAD models as ground truth. On the other hand, the Heidelberg datasets~\cite{7789500,meister2012outdoor} provide challenging and diverse cases for optical flow to assess algorithms robustness.

\subsection{Stereo}
A similiar trend can be observed for stereo datasets in \autoref{tab:stereo}. Once again, retrieving accurate ground truth for highly-dynamic scenes can turn out to be a challenge even with the use of LIDARs, as motion distortion can have considerable effect. Consequently, commonly used benchmarks are either static~\cite{988771} or synthetic~\cite{Butler:ECCV:2012}.

Stereo datasets are nonetheless useful in order to evaluate sensor fusion, especially since multi-modal learning is becoming more important. One interesting trend which can be observed in this table is the increase of resolution over time, which is also valid for every other tasks. This improvement in resolution embodies the hardware breakthroughs made over the years, but also begs the precision-performance question which future algorithms will have to face.

\subsection{Localization and mapping}
It also seems that localization/mapping datasets has not benefited from the same level of attention as object detection, which can be observed by the size of \autoref{tab:slam}. Most mapping and localization algorithms do not make use of deep-learning algorithms, meaning the need for large and diverse amount of data is not as critical. It can also be observed that localization and mapping datasets have been using LiDARs a lot earlier than those for object detection.

Centimeter precise measurements from DGPS or RTK are available for visual or point cloud odometry for most datasets. However, there is no datasets that provide ground truth for SLAM or mapping. Only qualitative evaluation and loop-closing can be used to evaluate the quality of a generated map.

\subsection{Behavior}
Finally, \autoref{tab:behavior} presents datasets which focus on behavioral aspects of driving. It can be noted that these kind of datasets have only recently started to gain attention. Also, since they are so recent, there is no clear annotation or methodology defined to quantify and capture driving behaviors. These datatsets are thus highly different from one another. It should also be noted that a recent trend tries to predict steering angle, brake or gas pedal, for use in \emph{end-to-end learning}.


\begin{landscape}

\begin{table}[ht]
\caption{Object Detection Datasets}
\begin{adjustbox}{max width=1.3\textheight}
\begin{tabular}{|l|l|c|c|l|cccc|ccccc|cccccc|cc|c|c|c|}
    \hline
    \multicolumn{1}{|c|}{\multirow{2}{*}{Name}} &
    \multicolumn{1}{c|}{\multirow{2}{*}{Provider}}  & 
    \multicolumn{1}{c|}{\multirow{2}{*}{Cit.}} &
    \multicolumn{1}{c|}{\multirow{2}{*}{Year}} &
    \multicolumn{1}{c|}{\multirow{2}{*}{Location}} &
    \multicolumn{4}{c|}{Diversity} & \multicolumn{5}{c|}{Traffic} &
    \multicolumn{6}{c|}{Sensors} &
    \multicolumn{2}{c|}{Annot.} &
    \multicolumn{1}{c|}{\multirow{2}{*}{Track.}} &
    \multicolumn{1}{c|}{Frames} &
    \multicolumn{1}{c|}{\multirow{2}{*}{Classes}} \\
    & & & & & R & S & N & D & U & H & Re & Ru & C & Vi & Li & GPS & Th & If & Ra & 2D & 3D & & (k) & \\
    \hline
    PandaSet \cite{panda_set}&  Hesai \& Scale & N/A & 2019 &  Silicon Valley & ? & ? & ? & ? & ? & ? & ? & ? & ? & X & X** & X &  &  &  & X & X &  & 60 & 28 \\
    nuScenes \cite{DBLP:journals/corr/abs-1903-11027} &  Aptiv & 16 & 2019 & Boston, Singapore & X &  & X & X & X &  &  &  &  & X & X & X &  &  & X & X & X & X & 40 & 23 \\
    Waymo Open \cite{waymo_open_dataset} &  Waymo Inc. & N/A & 2019 &  United States & X &  & X & X & X & X & X & X &  & X & X &  &  &  &  & X & X & X & 200 & 4/5 \\
    Lyft Level5 \cite{lyft2019}&  Lyft & N/A & 2019 & Palo Alto, London, Munich & ? & ? & ? & ? & ? & ? & ? & ? & ? & X & X & X &  &  &  & X & X & X & 55 & 23 \\
    H3D \cite{DBLP:journals/corr/abs-1903-01568}&  Honda Res. Inst. & 4 & 2019 &  San Francisco &  &  &  &  & X &  &  &  &  & X & X & X &  &  &  &  & X & X & 27 & 8 \\
    Boxy \cite{Boxy2019TBD}&  Bosch N.A. Res. & 0 & 2019 &  Unknown & X &  &  & X & X & X &  &  &  & X &  &  &  &  &  & X & X &  & 200 & 1 \\
    BLVD \cite{DBLP:journals/corr/abs-1903-06405}&  VCCIV & 2 & 2019 &  Changshu &  &  & X &  & X &  &  &  &  & X & X &  &  &  &  &  & X & X & 120 & 3 \\
    Road Damage \cite{DBLP:journals/corr/abs-1801-09454}&  University of Tokyo & 31 & 2018 &  Japan & X &  &  &  & X &  & X &  &  & X* &  &  &  &  &  & X &  &  & 9 & 8 \\
    KAIST Multispectral \cite{8293689}&  KAIST & 13 & 2018 &  Seoul &  &  & X & X & X &  & X &  & X & X & X & X & X &  &  & X &  & X & 8.9 & 6 \\
    BDD \cite{DBLP:journals/corr/abs-1805-04687}&  U. of California & 103 & 2018 & San Francisco, New York & X & X & X & X & X & X & X & X &  & X &  & X &  &  &  & X &  & X & 100 & 10 \\
    Apollo Open Platform \cite{apollo_open_platform} &  Baidu Inc. & N/A & 2018 &  China & X &  &  &  & X & X & X & X &  & X & X &  &  &  &  & X & X & X & 20 & 4/5 \\
    EuroCity \cite{braun2019eurocity} &  Delft U. of Tech. & 6 & 2018 &  European Cities & X &  & X & X & X &  &  &  &  & X &  &  &  &  &  & X &  & X & 47 & 7 \\
    FLIR Thermal Sensing \cite{flir_dataset} &  FLIR Sytems Inc. & N/A & 2018 &  Santa Barbara &  &  & X & X & X & X &  &  &  & X &  &  & X &  &  & X &  & X & 14 & 5 \\
    NightOwls \cite{Neumann18b} &  Oxford U. & 3 & 2018 & Germany,  Netherlands, UK & X & X & X & X & X &  & X &  &  & X &  &  &  &  &  & X &  & X & 279 & 4 \\
    TuSimple \cite{tusimple_dataset} &  TuSimple & N/A & 2017 &  Unknown & X &  &  &  &  & X &  &  &  & X &  &  &  &  &  & X &  & X & 12 & 1$\dagger$ \\
    NEXET \cite{nexar_dataset}&  Nexar & N/A & 2017 &  Around the globe & X & X & X &  & X & X & X & X &  & X &  &  &  &  &  & X &  &  & 55 & 5 \\
    Multi-spectral Object Detection \cite{Takumi:2017:MOD:3126686.3126727} &  U. of Tokyo & 5 & 2017 &  Tokyo &  &  & X & X &  &  &  &  & X & X &  &  & X &  &  & X &  &  & 7.5 & 5 \\
    Bosch Small Traffic Lights \cite{BehrendtNovak2017ICRA} &  Bosch N. A. Res. & 45 & 2017 &  San Francisco & X &  &  &  & X &  & X & X &  & X &  &  &  &  &  & X &  & X & 13 & 15 \\
    CityPersons \cite{DBLP:journals/corr/ZhangBS17} &  Max Planck Inst. (Info.) & 105 & 2017 & Germany, France, Switz. &  &  &  &  & X &  &  &  &  & X &  & X &  &  &  & X &  &  & 5 & 4 \\
    Udacity \cite{udacity_dataset} &  Udacity & N/A & 2016 &  Mountain View &  &  &  &  & X &  &  &  &  & X &  &  &  &  &  & X &  & X & 34 & 3 \\
    JAAD \cite{8265243} &  York University & 16 & 2016 & Ukraine, Canada & X & X & X & X & X &  &  &  &  & X &  &  &  &  &  & X &  & X & 88 & 1$\dagger\dagger$ \\
    Elektra (CVC-14) \cite{YSocarras:2013} &  Auton. U. of Barcelona & N/A & 2016 &  Barcelona &  &  & X &  & X &  &  &  &  & X &  &  &  & X &  & X &  & X & 10 & 1 \\
    Tsinghua-Daimler Cyclist \cite{7535515}&  Daimler AG & 41 & 2016 &  Beijing &  &  &  &  & X &  &  &  &  & X &  &  &  &  &  & X &  &  & 15 & 6 \\
    KAIST Multispectral Pedestrian \cite{7298706} &  KAIST & 195 & 2015 &  Seoul &  &  & X & X & X &  &  &  &  & X &  & X & X &  &  & X &  & X & 95 & 3 \\
    Highway Workzones \cite{6876163} &  CMU & 21 & 2015 &  United States & X &  &  &  &  & X &  &  &  & X &  &  &  &  &  & X &  & X & 17 & 9 \\
    KITTI \cite{DBLP:journals/corr/abs-1803-09719}&  Karlsruhe Inst. of Tech. & 5415 & 2013 &  Karlsruhe &  &  &  &  & X & X & X & X & X & X & X & X &  &  &  & X & X & X & 15 & 2 \\
    German Traffic Sign \cite{6033395}&  Ruhr U. & 298 & 2013 &  Germany & X &  &  & X & X & X & X & X &  & X* &  &  &  &  &  & X &  &  & 5 & 43 \\
    LISA Traffic Sign \cite{6335478}&  U. of California & 382 & 2012 &  San Diego &  &  & X & X & X &  &  &  &  & X* &  &  &  &  &  & X &  &  & 6.6 & 47 \\
    TME Motorway \cite{TMEMotorwayDataset}&  Czech Tech. U. & 127 & 2011 &  Northern Italy &  &  &  &  &  & X &  &  &  & X &  &  &  &  &  & X &  & X & 30 & 2 \\
    Belgian Traffic Sign \cite{5403121}&  ETH Zürich & 261 & 2011 &  Belgium &  &  &  &  & X &  &  &  &  & X* &  &  &  &  &  & X &  &  & 9 & 62 \\
    LISA Vehicle Detection \cite{5411825}& U. of California & 333 & 2010 &  California &  &  &  &  & X & X &  &  &  & X &  &  &  &  &  & X &  & X & 2.2 & 1 \\
    TUD-Brussels Pedestrian \cite{5206638} & Max Planck Inst. (Info) & 182 & 2009 &  Belgium &  &  &  &  & X &  &  &  &  & X &  &  &  &  &  & X &  & X & 1.6 & 1 \\
    ETH Pedestrian \cite{4587581}&  ETH Zürich & 547 & 2009 &  Zürich &  &  &  &  & X &  &  &  &  & X &  &  &  &  &  & X & X & X & 4.8 & 1 \\
    Caltech Pedestrian \cite{dollarCVPR09peds} &  Caltech & 1009 & 2009 &  Los Angeles &  &  &  &  & X &  &  &  &  & X &  &  &  &  &  & X &  & X & 250 & 1 \\
    Daimler Pedestrian \cite{4657363} &  Daimler AG & 1177 & 2008 &  Beijing &  &  &  &  & X &  &  &  &  & X &  &  &  &  &  & X &  & X & 21 & 1 \\
    \hline
    \multicolumn{25}{l}{R: Rain, S: Snow, N: Night, D: Dawn/Dusk, U: Urban, H: Highway, Re: Residential, Ru: Rural, C: Campus, Vi: Video, Li: LiDAR, Th: Thermal Camera, If: Infrared Camera, Ra: Radar} \\
    \multicolumn{25}{l}{*Non-continuous frames \quad\quad **Solid-state LiDAR \quad\quad $\dagger$also velocity \quad\quad $\dagger\dagger$also age, gender and direction} \\
\end{tabular}
\end{adjustbox}
\label{tab:object}
\end{table}


\begin{table}[ht]
\caption{Object Segmentation Datasets}
\begin{adjustbox}{max width=1.3\textheight}
\begin{tabular}{|l|l|c|c|l|cccc|ccccc|cccc|ccc|c|c|}
    \hline
    \multicolumn{1}{|c|}{\multirow{2}{*}{Name}} &
    \multicolumn{1}{c|}{\multirow{2}{*}{Provider}}  & 
    \multicolumn{1}{c|}{\multirow{2}{*}{Cit.}} &
    \multicolumn{1}{c|}{\multirow{2}{*}{Year}} &
    \multicolumn{1}{c|}{\multirow{2}{*}{Location}} &
    \multicolumn{4}{c|}{Diversity} & \multicolumn{5}{c|}{Traffic} &
    \multicolumn{4}{c|}{Sensors} &
    \multicolumn{3}{c|}{Annot.} &
    \multicolumn{1}{c|}{Frames} &
    \multicolumn{1}{c|}{\multirow{2}{*}{Classes}} \\
    & & & & & R & S & N & D & U & H & Re & Ru & C & Vi & Li & GPS & Th & Px & In & Pc & (k) & \\
    \hline
    PandaSet \cite{panda_set}&  Hesai \& Scale & N/A & 2019 &  Silicon Valley & ? & ? & ? & ? & ? & ? & ? & ? & ? & X & X** & X &  & X & ? & X & 60 k & 37 \\
    SemanticKitti \cite{DBLP:journals/corr/abs-1904-01416} &  University of Bonn & 1 & 2019 &  Karlsruhe &  &  &  &  & X &  & X &  &  &  & X & X &  &  &  & X & 43 k & 28 \\
    Highway Driving \cite{Kim2018HighwayDD} &  Korea Advanced Institute of Science and Technology & 1 & 2019 &  Unknown &  &  &  &  &  & X &  &  &  & X &  &  &  & X &  &  & 1.2 k & 10 \\
    ApolloScape \cite{DBLP:journals/corr/abs-1803-06184}&  Baidu Inc. & 7 & 2018 &  China & X & X & X &  & X & X & X & X &  & X & X & X &  & X & X & X & 147 k & 35 \\
    Apollo Open Platform \cite{apollo_open_platform}&  Baidu Inc. & N/A & 2018 &  China & X &  &  &  & X & X & X & X &  & X &  &  &  & X &  &  & 17 k & 26 \\
    BDD \cite{DBLP:journals/corr/abs-1805-04687}&  U. of California & 103 & 2018 & San Francisco, New York & X & X & X & X & X & X & X & X &  & X &  & X &  & X & X &  & 10 k & 40 \\
    Wilddash \cite{Zendel_2018_ECCV}&  Austrian Inst. of Tech. & 8 & 2018 &  Around the globe & X & X & X & X & X & X & X & X &  & X &  &  &  & X & X &  & 226 & 30 \\
    Raincouver \cite{7970170}&  U. of British Columbia & 3 & 2018 &  Vancouver & X & X & X & X & X & X & X & X &  & X &  &  &  & X &  &  & 326 & 3 \\
    MightyAi Sample \cite{mighty_ai_dataset}&  MightyAI & N/A & 2018 &  Seattle &  &  &  &  & X & X & X &  &  & X &  &  &  & X & X &  & 200 & 41 \\
    IDD \cite{DBLP:journals/corr/abs-1811-10200}&  IIIT Hyderabad & 6 & 2018 & Hyderabad,  Bangalore &  &  &  & X & X &  & X & X &  & X &  &  &  & X &  &  & 10 k & 34 \\
    Mapillary Vistas \cite{MVD2017} &  Mapillary AB & 139 & 2017 &  Around the globe & X & X & X & X & X & X & X & X &  & X* &  &  &  & X & X &  & 20 k & 66 \\
    Multi-Spectral Semantic Segmentation \cite{8206396} &  U. of Tokyo & 11 & 2017 &  Tokyo &  &  & X & X & X &  &  &  &  & X &  &  & X & X &  &  & 1.6 k & 8 \\
    Cityscapes \cite{Cordts2016Cityscapes}&  Daimler AG & 1729 & 2016 & Germany,  France, Switzerland &  &  &  &  & X &  &  &  &  & X &  & X &  & X & X &  & 5 k & 30 \\
    LostAndFound \cite{DBLP:journals/corr/PinggeraRGFRM16}&  Daimler AG & 14 & 2016 &  Germany &  &  &  &  & X & X & X &  &  & X* &  &  &  &  & X &  & 2 k & 37 \\
    Daimler Urban Segmentation \cite{10.1007/978-3-642-40602-7_46}&  Daimler AG & 67 & 2014 &  Heidelberg &  &  &  &  & X &  &  &  &  & X &  &  &  & X &  &  & 500 & 5 \\
    KITTI Semantics \cite{DBLP:journals/corr/abs-1803-09719} &  Karlsruhe Institute of Technology & 5415 & 2013 &  Karlsruhe &  &  &  &  & X &  & X &  &  & X & X & X &  & X & X &  & 200 & 34 \\
    Standford Track \cite{Teichman2011}&  Stanford University & 146 & 2011 &  California &  &  &  &  & X &  &  &  & X &  & X & X &  &  & X & X & 14 k & 3 \\
    CamVid \cite{BrostowFC:PRL2008} &  University of Cambridge & 475 & 2009 &  Cambridge &  &  &  &  & X &  &  &  &  & X &  &  &  & X &  &  & 700 & 32 \\
    Daimler Pedestrian \cite{4657363} &  Daimler AG & 1177 & 2008 &  Beijing &  &  &  &  & X &  &  &  &  &  &  &  &  & X &  &  & 785 & 1 \\ 
    \hline
    \multicolumn{23}{l}{R: Rain, S: Snow, N: Night, D: Dawn/Dusk, U: Urban, H: Highway, Re: Residential, Ru: Rural, C: Campus, Vi: Video, Li: LiDAR, Th: Thermal Camera, Px: Pixel, In: Instance, Pc: Point Cloud} \\
    \multicolumn{23}{l}{*Non-continuous frames \quad\quad **Solid-state LiDAR} \\
\end{tabular}
\end{adjustbox}
\label{tab:segementation}
\end{table}


\begin{table}[ht]
\caption{Lane Detection Datasets}
\begin{adjustbox}{max width=1.3\textheight}
\begin{tabular}{|l|l|c|c|l|cccc|ccccc|ccc|cccc|c|c|}
    \hline
    \multicolumn{1}{|c|}{\multirow{2}{*}{Name}} &
    \multicolumn{1}{c|}{\multirow{2}{*}{Provider}}  & 
    \multicolumn{1}{c|}{\multirow{2}{*}{Cit.}} &
    \multicolumn{1}{c|}{\multirow{2}{*}{Year}} &
    \multicolumn{1}{c|}{\multirow{2}{*}{Location}} &
    \multicolumn{4}{c|}{Diversity} & \multicolumn{5}{c|}{Traffic} &
    \multicolumn{3}{c|}{Sensors} &
    \multicolumn{4}{c|}{Annot.} &
    \multicolumn{1}{c|}{\multirow{2}{*}{Frames}} &
    \multicolumn{1}{c|}{\multirow{2}{*}{Classes}} \\
    & & & & & R & S & N & D & U & H & Re & Ru & C & Vi & Li & GPS & Px & Sp & Bx & Pc & & \\
    \hline
    Unsupervised Llamas \cite{Llamas2019}&  Bosch N. A. Research & 0 & 2019 &  California &  &  &  &  &  & X &  &  &  & X & X & X & X &  &  & X & 100 k & 5 \\
    BDD \cite{DBLP:journals/corr/abs-1805-04687} &  U. of California & 103 & 2018 & San Francisco, New York & X & X & X & X & X & X & X & X &  & X &  & X &  & X &  &  & 80 k & 11 \\
    ApolloScape \cite{DBLP:journals/corr/abs-1803-06184}&  Baidu Inc. & 7 & 2018 &  China & X & X & X &  & X & X & X & X &  & X & X & X & X &  &  & X & 170 k & 27 \\
    CULane \cite{pan2018SCNN} &  University of Hong Kong & 42 & 2018 &  Beijing &  &  & X &  & X & X & X & X &  &  &  &  &  & X &  &  & 130 k & 9 \\
    VPGNet \cite{DBLP:journals/corr/abs-1710-06288}&  KAIST & 44 & 2017 &  Seoul & X &  & X &  & X & X &  &  &  & X &  & X &  & X &  &  & 21 k & 17 \\
    TuSimple \cite{tusimple_dataset}&  TuSimple & N/A & 2017 &  Unknown & X &  &  &  &  & X &  &  &  & X &  &  &  & X &  &  & 6.4 k & 4 \\
    TRoM \cite{8317749} &  Tsinghua University & 1 & 2017 &  Beijing & X &  &  & X & X & X &  &  &  & X &  &  &  & X &  &  & 712 & 19 \\
    KITTI Road  \cite{DBLP:journals/corr/abs-1803-09719} &  Karlsruhe Institute of Technology & 5415 & 2013 &  Karlsruhe &  &  &  &  & X &  &  &  &  & X & X & X & X &  &  &  & 579 & 2 \\
    Road Marking \cite{6232144} &  Honda Research Institute & 68 & 2012 &  California &  &  & X & X & X & X &  &  &  & X &  &  &  &  & X &  & 1.4 k & 23 \\
    CalTech Lanes \cite{DBLP:journals/corr/Aly14}&  California Institute of Technology & 577 & 2008 & Washington, Cordova &  &  &  &  & X & X &  &  &  & X &  &  &  & X &  &  & 1.2 k & 2 \\
    \hline
    \multicolumn{23}{l}{R: Rain, S: Snow, N: Night, D: Dawn/Dusk, U: Urban, H: Highway, Re: Residential, Ru: Rural, C: Campus, Vi: Video, Li: LiDAR, Th: Thermal Camera, Px: Pixel,Sp: Spline, Bx: Bounding Box, Pc: Point Cloud}
\end{tabular}
\end{adjustbox}
\label{tab:lane}
\end{table}


\begin{table}[ht]
\caption{Optical Flow Datasets}
\begin{adjustbox}{max width=1.3\textheight}
\begin{tabular}{|l|l|c|c|l|cccc|ccc|cc|c|c|}
    \hline
    \multicolumn{1}{|c|}{\multirow{2}{*}{Name}} &
    \multicolumn{1}{c|}{\multirow{2}{*}{Provider}}  & 
    \multicolumn{1}{c|}{\multirow{2}{*}{Cit.}} &
    \multicolumn{1}{c|}{\multirow{2}{*}{Year}} &
    \multicolumn{1}{c|}{\multirow{2}{*}{Location}} &
    \multicolumn{4}{c|}{Diversity} & \multicolumn{3}{c|}{Traffic} &
    \multicolumn{2}{c|}{Annot.} &
    \multicolumn{1}{c|}{\multirow{2}{*}{Ground Truth Retrieval}} &
    \multicolumn{1}{c|}{\multirow{2}{*}{Frames}} \\
    & & & & & R & S & N & D & U & H & Ru & Scene Flow & Optical Flow &  & \\
    \hline
    HD1K \cite{7789500} &  Heidelberg Collaboratory, Bosch & 20 & 2016 &  Heidelberg & X &  & X & X & X &  &  &  & X & Stereo/Lidar Fusion + ICP & 1000 \\
    KITTI Flow 2015 \cite{7298925}&  Karlsruhe Institute of Technology & 629 & 2015 &  Karlsruhe &  &  &  &  & X &  &  & X & X & Stereo/Lidar Fusion + 3D CAD Models & 400 \\
    KITTI Flow 2012 \cite{6248074} &  Karlsruhe Institute of Technology & 3508 & 2012 &  Karlsruhe &  &  &  &  & X &  &  &  & X & Stereo/Lidar Fusion + ICP & 391 \\
    HCI Challenging Stereo \cite{meister2012outdoor} &  Heidelberg Collaboratory, Bosch & 57 & 2012 &  Hildesheim & X &  & X & X & X & X &  &  & X & Stereo + Monoscopic Tracking & 10 k\\
    \hline
    \multicolumn{16}{l}{R: Rain, S: Snow, N: Night, D: Dawn/Dusk, U: Urban, H: Highway, Ru: Rural}
\end{tabular}
\end{adjustbox}
\label{tab:optical}
\end{table}


\begin{table}[ht]
\caption{Stereo Datasets}
\begin{adjustbox}{max width=1.3\textheight}
\begin{tabular}{|l|l|c|c|c|c|c|c|ccccc|c|}
    \hline
    \multicolumn{1}{|c|}{\multirow{2}{*}{Name}} &
    \multicolumn{1}{c|}{\multirow{2}{*}{Provider}}  & 
    \multicolumn{1}{c|}{\multirow{2}{*}{Cit.}} &
    \multicolumn{1}{c|}{\multirow{2}{*}{Year}} &
    \multicolumn{1}{c|}{\multirow{2}{*}{Location}} &
    \multicolumn{1}{c|}{\multirow{2}{*}{Resolution}} &
    \multicolumn{1}{c|}{\multirow{2}{*}{Baseline}} &    \multicolumn{1}{c|}{\multirow{2}{*}{Frames}} &
    \multicolumn{5}{c|}{Diversity} &
    \multicolumn{1}{c|}{\multirow{2}{*}{Ground Truth}} \\
     & & & & & & & & Env & We & Il & Ti & Se &  \\
    \hline
    Complex Urban \cite{jeong2019complex} &  KAIST & 14 & 2018 & South Korea & 1600x1200 & 47 & 36 k & X &  & X &  &  & LiDAR \\
    ApolloScape \cite{DBLP:journals/corr/abs-1803-06184} &  Baidu Inc. & 7 & 2018 &  China & 3384x2710 & 30 & 5 k & X &  &  &  &  & Stereo/Lidar Fusion + 3D CAD Models \\
    HD1K \cite{7789500} &  Heidelberg Collaboratory, Bosch & 20 & 2016 &  Heidelberg & 1080x2560 & 30 & 10 k &  & X & X &  &  & Stereo/Lidar Fusion + 3D CAD Models \\
    Elektra (CVC-13) \cite{6220232}&  Autonomous University of Barcelona & 73 & 2016 &  Barcelona & 640x480 & 12 & 110 &  & X & X &  &  & Manual + 3D CAD Models \\
    CCSAD \cite{inproceedingsCCSAD} &  Centro de Investigación En Matemáticas & 7 & 2015 &  Mexico & 1096x822 & 50 & 96 k & X &  & X &  &  & - \\
    KITTI Stereo 2015 \cite{7298925}&  Karlsruhe Institute of Technology & 629 & 2015 &  Karlsruhe & 1240x376 & 54 & 396 &  &  & X &  &  & Stereo/Lidar Fusion + 3D CAD Models \\
    Oxford RobotCar \cite{RobotCarDatasetIJRR} &  Oxford University & 246 & 2015 &  Central Oxford & 1280x960 & 24 & 1 M + &  & X & X & X & X & LiDAR \\
    Málaga Stereo and Urban \cite{BlancoClaraco2014TheMU} &  University of Málaga & 107 & 2013 &  Málaga & 1024x768 & 12 & 111 k &  &  & X &  &  & LiDAR \\
    Stixel \cite{DBLP:journals/corr/CordtsRSPERPF17} &  Daimler AG & 91 & 2013 &  Germany & 1024x333 & 21 & 2500 &  & X &  &  &  & Manual \\
    KITTI Stereo 2012 \cite{6248074}&  Karlsruhe Institute of Technology & 3508 & 2012 &  Karlsruhe & 1242x374 & 54 & 400 &  &  & X &  &  & Stereo/Lidar Fusion + ICP \\
    Ladicky \cite{Ladicky2012} &  Oxford Brookes University & 91 & 2012 & Leuven & 316x256 & 150 & 70 &  &  &  &  &  & Manual \\
    \hline
    \multicolumn{14}{l}{Env: Environment, We: Weather, Il: Illumination, Ti: Daytime, Se: Seasons} \\
\end{tabular}
\end{adjustbox}
\label{tab:stereo}
\end{table}


\begin{table}[ht]
\caption{Localization and Mapping Datasets}
\begin{adjustbox}{max width=1.3\textheight}
\begin{tabular}{|l|l|c|c|c|c|c|cccccc|cc|c|}
    \hline
    \multicolumn{1}{|c|}{\multirow{2}{*}{Name}} &
    \multicolumn{1}{c|}{\multirow{2}{*}{Provider}}  & 
    \multicolumn{1}{c|}{\multirow{2}{*}{Cit.}} &
    \multicolumn{1}{c|}{\multirow{2}{*}{Year}} &
    \multicolumn{1}{c|}{\multirow{2}{*}{Location}} &
    \multicolumn{1}{c|}{Length} &
    \multicolumn{1}{c|}{Frames} &
    \multicolumn{6}{c|}{Sensor} &
    \multicolumn{2}{c|}{Ground Truth} &
    \multicolumn{1}{c|}{\multirow{2}{*}{Distinctiveness}} \\
     & & & & & (km) & (k) & Vi & 2DLi & 3DLi & Ve & If & Ev & Pose & Map &  \\
    \hline
    StreetLearn \cite{DBLP:journals/corr/abs-1903-01292}&  DeepMind & 3 & 2019 & New York, Pittsburgh & 1100.0 & 114 & X &  &  &  &  &  & GPS & X & Collection of Google Street View panoramas \\
    UTBM RoboCar \cite{utbm_robocar_dataset}&  U. of Tech. of Belfort-Montbéliard & N/A & 2019 &  Montbéliard & 63.4 & 220 & X & X & X &  &  &  & RTK &  & Long-term, day/night diversity, seasons \\
    Multi Vehicle Stereo Event Camera \cite{8288670}& U. of Pennsylvania & 29 & 2018 &  Pennsylvania & 9.6 & 36 & X &  & X &  &  & X & GPS/IMU & X & Event camera, map as ground truth \\
    Complex Urban \cite{jeong2019complex} &  KAIST & 14 & 2018 &  South Korea & 44.8 & - &  & X & X &  &  &  & RTK & X & Focus on diverse and complex urban scenarios \\
    Apollo Open Platform \cite{apollo_open_platform}&  Baidu Inc. & N/A & 2018 &  China & 3.0 & - &  &  & X &  &  &  & RTK &  & Multi-Sensor Localization \\
    comma2k19 \cite{1812.05752}&  comma.ai & 1 & 2018 & San Francisco, San Jose & 2000 + & 2400 & X &  &  &  &  &  & GNSS &  & Raw GNSS measurements, long-term \\
    Oxford RobotCar \cite{RobotCarDatasetIJRR}&  Oxford University & 246 & 2015 &  Central Oxford & 1 and 010.5 & 1000 + & X & X & X &  &  &  & DGPS &  & Long-Term, same segment, weather diversity \\
    AMUSE \cite{6595954}&  Linköping University & 12 & 2013 &  Linköping & 24.4 & 108 & X &  &  & X &  &  & GPS/IMU &  & Velocity and height sensors and snow \\
    KITTI Odometry \cite{DBLP:journals/corr/abs-1803-09719}&  Karlsruhe Inst. of Tech. & 5415 & 2013 &  Karlsruhe & 39.2 & 41 & X &  & X &  &  &  & RTK &  & Widely-used benchmark for SLAM \\
    Málaga Stereo and Urban \cite{BlancoClaraco2014TheMU}&  University of Málaga & 107 & 2013 &  Málaga & 36.8 & 111 & X & X &  &  &  &  & RTK &  & Direct sun \\
    Cheddar Gorge \cite{articleCheddar} &  BAE Systems & 13 & 2011 &  Cheddar Gorge & 31.8 & 86 & X &  & X &  & X &  & DGPS &  & Infrared camera and  Countryside mapping \\
    The annotated laser data set \cite{doi:10.1177/0278364910389840}&  CMU & 11 & 2011 &  Pittsburgh & 5.0 & 5 & X & X &  &  &  &  & - &  & 2D Lidar obstacle detection and SLAM \\
    Ford \cite{articleFord}&  U. of Michigan & 177 & 2010 &  Michigan & 5.1 & 7 & X & X & X &  &  &  & DGPS &  & Omni cameras, precise camera-laser registration \\
    \hline
    \multicolumn{16}{l}{Vi: Video, 2DLi: 2D LiDAR, 3DLi, 3D LiDAR, Ve: Velocity Sensor, If: Infrared Camera, Ev: Event Camera}

\end{tabular}
\end{adjustbox}
\label{tab:slam}
\end{table}

    
\begin{table}[ht]
\caption{Behavior Datasets}
\begin{adjustbox}{max width=1.3\textheight}
\begin{tabular}{|l|l|l|l|l|l|cccc|ccc|ccccccc|c|c|}
    \hline
    \multicolumn{1}{|c|}{\multirow{2}{*}{Name}} &
    \multicolumn{1}{c|}{\multirow{2}{*}{Provider}}  & 
    \multicolumn{1}{c|}{\multirow{2}{*}{Cit.}} &
    \multicolumn{1}{c|}{\multirow{2}{*}{Year}} &
    \multicolumn{1}{c|}{\multirow{2}{*}{Location}} &
    \multicolumn{1}{c|}{\multirow{2}{*}{Volume}} &
    \multicolumn{4}{c|}{Diversity} &
    \multicolumn{3}{c|}{Traffic} &
    \multicolumn{7}{c|}{Sensor} &
    \multicolumn{1}{c|}{Annotation} &
    \multicolumn{1}{c|}{Objective}\\
     & & & & & & R & S & N & D & U & H & Re & Vi & Fc & Li & GPS & IMU & CAN & Ev & &  \\
    \hline
    DBNet \cite{DBNet2018} & Shangai Jiao Tong University&4&2018& China&17&X&X&X&X&X&X&X&X&&X&X&X&X&&-&Steering/Velocity Prediction \\
    HDD \cite{DBLP:journals/corr/abs-1811-02307}& Honda Research Institute&23&2018& San Francisco&104&&&&&X&X&X&X&&X&X&X&X&&Maneuvers&Maneuver Prediction, Causal Reasoning \\
    comma2k19 \cite{1812.05752}& comma.ai&1&2018& San Francisco, San Jose&35&&&&&&X&&X&&&X&X&X&&-&Steering Prediction \\
    DDD17 \cite{DBLP:journals/corr/abs-1711-01458}& ETH Zürich&31&2017& Switzerland, Germany&12&X&&X&X&X&X&&X&&&&&X&X&-&Steering Prediction \\
    Brain4Cars \cite{DBLP:journals/corr/JainKSRSS16}& Cornell University&35&2016& United States&22&&&&&X&X&&X&X&&X&&X&&Maneuvers, Street Maps&Maneuver Prediction \\
    Dr(eye)ve \cite{DBLP:journals/corr/PalazziACSC17}& University of Modena&24&2016& Modena&6&X&&X&&X&X&X&X&&&X&X&&&Gaze Maps, Pupil Dilation&Attention Monitoring \\
    JAAD \cite{8265243}& York University &16&2016& Ukraine, Canada &240&X&X&X&X&X&&&X&&&&&&&Pedestrian/Driver Behavior, Context&Pedestrian/Driver Behavior \\
    UAH \cite{7795584}& University of Alcalá&29&2016& Madrid&8&&&&&&X&&X&&&X&X&&&Driver Behavior, Maneuvers&Driver Behavior, Maneuver Prediction \\
    Elektra  (DrivFace) \cite{Diaz-Chito:2016:RFS:2951121.2951316}& Autonomous University of Barcelona &10&2016& Barcelona &-&&&&&X&&&&X&&&&&&Gaze Direction, Facial Keypoints&Gaze Monitoring \\
    Udacity \cite{udacity_dataset}& Udacity &N/A&2016& Mountain View &8&&&&&X&X&&X&&X&X&X&&&-&Steering Prediction \\
    comma.ai \cite{DBLP:journals/corr/SantanaH16}& comma.ai&90&2016& San Francisco&7&&&&&&X&&X&&&X&X&X&&-&Steering Prediction \\
    DIPLECS Surrey \cite{7293673}& University of Surrey&28&2015& Surrey&0.5&&&&&X&X&X&X&&&&&X&&-&Steering Prediction \\
    DIPLECS Sweden \cite{7293673}& Autoliv&36&2015& Stockholm&3&&&&&X&X&&X&&&X&&&&Maneuvers, Context&Maneuver Prediction \\
    EISATS (Set 11) \cite{inproceedingsEISATS}& University of Auckland&5&2010& Germany, New Zealand&<1&&&&&X&&&X&&&&&&&-&Gaze Monitoring \\
    \hline
    \multicolumn{22}{l}{R: Rain, S: Snow, N: Night, D: Dawn/Dusk, U: Urban, H: Highway, Re: Residential, Vi: Video, Fc: Face Camera, Li: LiDAR, CAN: Controller Area Network Data, Ev: event-camera} \\
\end{tabular}
\end{adjustbox}
\label{tab:behavior}
\end{table}

\end{landscape}

\end{spacing}